# Beyond Automation: Socratic AI, Epistemic Agency, and the Implications of the Emergence of Orchestrated Multi-Agent Learning Architectures

Peer-Benedikt Degen[1] & Igor Asanov[2]

1: Department of Empirical Educational Research, Institute of Educational Science, Faculty of Human Sciences, University of Kassel, Nora-Platiel-Straße 1, 34127, Kassel, Germany

2: International Centre for Higher Education Research (INCHER), University of Kassel, Mönchebergstraße 17, 34125, Kassel, Germany

## Author note

Peer-Benedikt Degen  https://orcid.org/0000-0002-4330-0893

Igor Asanov  https://orcid.org/0000-0002-8091-4130

We have no known conflict of interest to disclose. Correspondence concerning this article should be addressed to Ben Degen, University of Kassel, Mönchebergstr. 21a, 34125 Kassel, Germany. Email: Degen@uni-kassel.de

## CRediT author statement

**Peer-Benedikt Degen:** Conceptualization, Methodology, Investigation, Visualization, Data curation, Formal analysis, Writing – Original Draft. **Igor Asanov:** Formal analysis, Writing – Review & Editing

## Acknowledgments

We thank Dr. Marit Kastaun and Lars Meyer-Odewald for granting access to the courses employed in the experimental implementation. We are also grateful to Prof. Dr. Dirk Krüger and Prof. Dr. Annette Upmeier zu Belzen for kindly providing the validated Ko-WADiS items used to assess research question competence in the biological domain. Finally, we thank Dr. Jane Southworth for her permission to reprint Fig. 4.

## Abstract

Generative AI is no longer a peripheral tool in higher education. It is rapidly evolving into a general-purpose infrastructure that reshapes how knowledge is generated, mediated, and validated. This paper presents findings from a controlled experiment evaluating a Socratic AI Tutor, a large language model designed to scaffold student research question development through structured dialogue grounded in constructivist theory.

Conducted with 65 pre-service teacher students in Germany, the study compares interaction with the Socratic Tutor to engagement with an uninstructed AI chatbot. Students using the Socratic Tutor reported significantly greater support for critical, independent, and reflective thinking, suggesting that dialogic AI can stimulate metacognitive engagement and challenging recent narratives of de-skilling due to generative AI usage.

These findings serve as a proof of concept for a broader pedagogical shift: the use of multi-agent systems (MAS) composed of specialised AI agents. To conceptualise this, we introduce the notion of *orchestrated* MAS, modular, pedagogically aligned agent constellations, curated by educators, that support diverse learning trajectories through differentiated roles and coordinated interaction. To anchor this shift, we propose an adapted *offer-and-use model*, in which students appropriate instructional offers from these agents.

Beyond technical feasibility, we examine system-level implications for higher education institutions and students, including funding necessities, changes to faculty roles, curriculars, competencies and assessment practices. We conclude with a comparative cost-effectiveness analysis highlighting the scalability of such systems. In sum, this study contributes both empirical evidence and a conceptual roadmap for hybrid learning ecosystems that embed human–AI co-agency and pedagogical alignment.

# 1. Introduction

Generative artificial intelligence (GenAI) is no longer a peripheral tool in higher education. It is increasingly embedded into core academic practices, influencing how knowledge is generated, mediated, and validated (Chen, 2025; Menekse et al., 2025). While much attention has focused on GenAI's capacity to automate routine or administrative tasks, its role in shaping deeper cognitive and epistemic processes is only beginning to be explored as AI systems begin to exceed individual capabilities of humans (Haase & Hanel, 2023; Luo et al., 2025).

This paper examines the educational potential of GenAI not as a tool for offloading cognitive work, but as a dialogic partner that supports students' ability to think critically and formulate meaningful research questions. We report findings from a controlled experiment to assess feasibility and compare two AI-supported learning conditions: a Socratic AI Tutor, a large language model instructed to scaffold research question formulation through structured questioning, and a general-purpose, uninstructed AI chatbot. Conducted with pre-service teacher education students, the study included baseline assessments and two transfer tasks, enabling evaluation of both immediate learning gains and the application of inquiry strategies in novel contexts.

The results suggest that students who interacted with the Socratic AI Tutor perceived significantly greater support for critical, independent, and reflective thinking compared to those using the uninstructed AI chatbot. These findings indicate that dialogically structured AI systems can prompt metacognitive engagement and stimulate higher-order thinking processes. Rather than merely delivering information, the Socratic AI Tutor activated learners' capacity to question, reflect, and take ownership of their reasoning, supporting epistemic agency (Miller et al., 2018).

Based on these results, we propose that the future of AI in education lies not in single-purpose systems, but in orchestrated multi-agent architectures that challenge the industrial-era one-teacher-many-students paradigm and opens pathways for scalable, adaptive pedagogies grounded in shared inquiry: modular constellations of AI agents, each supporting distinct aspects of the learning process. This vision reimagines the structure of academic support and challenges traditional assumptions about the division of pedagogical labour.

In what follows, we outline the theoretical foundations of the Socratic AI Tutor, describe the design and methodology of the study, present empirical findings, and explore broader implications for curriculum, assessment, faculty roles, and institutional infrastructure. Our aim is to contribute to an emerging conversation on how generative AI can underpin new systems of learning that support inquiry, reflection, and creativity at scale.

## 2. Theoretical Foundations: Epistemic Agency, Socratic Questioning and Constructive Principles Against the Background of GenAI

Epistemic agency refers to learners' capacity to take responsibility for and exert control over the construction of knowledge (Odden et al., 2023). This involves a shift from learning-as-consumption to intentional knowledge advancement, where learners act as participants in a collective endeavour to improve shared ideas through inquiry, judgement, and self-regulated engagement with complex problems (Katsara & De Witte, 2019; Scardamalia & Bereiter, 1991). Rather than simply acquiring or applying existing knowledge, students acting with epistemic agency define their own questions, pursue open-ended investigations, and evaluate the adequacy of explanations through disciplinary practices.

When nurtured effectively, epistemic agency enables students to inhabit the role of epistemic agents: individuals or communities granted cognitive authority to shape what counts as knowledge within a domain (Miller et al., 2018; Stroupe, 2014). This orientation is increasingly vital in an era where generative AI systems pose both a challenge and an opportunity for engaging students in sustained, meaningful knowledge work.

However, the widespread adoption of generative AI in education also introduces dynamics that can work against the nurturing of epistemic agency. Many AI-powered tools are optimised for efficiency, fluency, and superficial correctness, producing seemingly authoritative responses that may discourage students from engaging in deeper inquiry (Bastani et al., 2024). By delivering polished answers with minimal cognitive effort, such systems risk reinforcing passive modes of knowledge consumption, displacing the need for question formulation, critical evaluation, or engagement with epistemic struggle and desirable difficulties that are essential for deeper learning (Blasco & Charisi, 2024; Groothuijsen et al., 2024).

One illustrative case is the use of recommender systems in educational contexts. Liu et al. (2023) developed CoQuest, an LLM-based agent designed to support students in formulating research questions by automatically proposing potential inquiries for them to pursue. While the system was intended to assist learners, it raises concerns about its effectiveness in actually fostering students' research competencies as it creates overreliance on AI-generated answers. This tendency toward overreliance is difficult to mitigate. In a study by Buçinca et al. (2021), attempts to reduce user dependence on AI-generated recommendations by providing explanatory rationales proved largely ineffective. The authors found that even when explanations were offered, users continued to make poorer decisions when the AI's suggestions were flawed, compared to scenarios in which no AI assistance was given at all (p. 188:2).

When learners become habituated to outsourcing thinking to AI, the development of reflective judgement and disciplinary reasoning may be compromised. Moreover, without explicit scaffolding, students may accept AI-generated outputs uncritically as has been shown above, mistaking plausibility for validity. In this context, it becomes essential to reorient AI's pedagogical function not as a provider of answers, but as a co-inquirer that helps sustain and support epistemic work.

## 2.1. Socratic Questioning

It is in this context that the Socratic method, or more precisely Socratic questioning becomes pedagogically significant. Socratic questioning represents a longstanding educational tradition, originating in classical antiquity with Socrates himself (Heckmann & Krohn, 2018), and has since informed a range of dialogic instructional approaches that share its core features (Knezic et al., 2010). At its core, it denotes a deliberate pedagogical process in which a teacher guides a student through structured questioning in order to foster the development of specific understandings or insights (Burbules & Bruce, 2001).

Despite debates regarding its precise definition and historical fidelity (Burbules & Bruce, 2001; Carey & Mullan, 2004)), contemporary educational interpretations converge on several key features: First, the presence of a facilitator who guides learners through intentional questioning, second a shift from adversarial cross-examination toward collaborative inquiry; and third a focus on deepening understanding rather than transmitting content. (Fahrner & Wolf, 2020; Katsara & De Witte, 2019; Knezic et al., 2010; Overholser, 1993).

Paul (1990, pp. 276-278) for example proposed a taxonomy of Socratic questions comprising six categories (see Table 1), each illustrated with representative question types. A later revision of this framework, published by Paul and Elder in 2007, offers a broader, but arguably more generalised, version that readers may also wish to consult.

**Table 1**

*Taxonomy of Socratic Questions including selected examples of questions.*

| Category | Selected examples |
| --- | --- |
| Questions of clarification | What do you mean by ________? Could you give me an example? Could you explain that further? Could you put that another way? How does this relate to our discussion (problem, issue)? |
| Questions that probe assumptions | What are you assuming? What could we assume instead? You seem to be assuming. How would |

| Category | Selected examples |
|---|---|
| Questions that probe reason and evidence | you justify taking this for granted? Is it always the case? What would be an example? Are these reasons adequate? Do you have any evidence for that? How does that apply to this case? But is that good evidence to believe that? |
| Questions about viewpoints or perspectives | You seem to be approaching this issue from perspective. Why have you chosen this rather than that perspective? How would other groups/types of people respond? Why? What would influence them? What would someone who disagrees say? |
| Questions that probe implications or consequences | What are you implying by that? What effect would that have? Would that necessarily happen or only probably happen? What is an alternative? |
| Questions about the question | How can we find out? What does this question assume? Why is this question important? Can we break this question down at all? Is the question clear? Do we understand it? Is this question easy or hard to answer? Why? |

*Note.* By Paul (1990, p. 276-278) as in [anonymised for review]

As can be seen, the questions posed are typically open-ended, recursive, prompt clarification, test assumptions, explore perspectives, and consider implications. In this way, the Socratic method aligns closely with constructivist learning theories that emphasise active knowledge construction through dialogue and reflection.

### 2.2. Constructivist Foundations

This form of dialogic scaffolding aligns closely with foundational principles of constructivist learning theory, which views knowledge as actively constructed rather than passively received. One of the central tenets in this tradition is Vygotsky's concept of the Zone of Proximal Development (ZPD). According to (Vygotsky, 1980), learners develop most effectively when engaging in tasks just beyond their independent capabilities, provided they receive appropriate guidance from a more *knowledgeable other*, traditionally a teacher, peer, or expert. This interaction creates a space in which learners are encouraged to extend their thinking, test ideas, and engage in increasingly complex reasoning.

Along similar lines, (Bruner, 1966) articulated the concept of the spiral curriculum, in which key ideas are revisited across time at progressively more complex levels. In this framework, learning is not linear but recursive and advances through cycles of exploration, reflection, and elaboration. Structured dialogue plays a central role in this process by providing the cognitive scaffolds necessary for learners to revisit earlier insights, challenge assumptions, and construct deeper conceptual frameworks over time.

Constructivist theory thus offers a robust foundation for dialogic pedagogies, such as the Socratic questioning, that can be used to foster epistemic agency through sustained, inquiry-driven interaction.

## 3. Design Principles of the Socratic AI Tutor

The design of the Socratic AI Tutor was guided by a singular pedagogical ambition: to scaffold student research question development without automating thought. Rather than positioning the AI as a source of information, we configured it as a dialogic partner capable of prompting, challenging, and refining learner thinking in real-time.

Acknowledging concerns over proprietary models (Palmer et al., 2024) and given that earlier fine-tuning attempts proved unsuccessful, as detailed in the study protocol [anonymised for review], we deployed OpenAI's GPT-4o model at the state of the art model at that time for its robust dialogue coherence and context sensitivity. To ensure logical consistency in the model's responses, the temperature setting, which controls the degree of randomness in word selection on a scale from 0 to 1, was set to 0.10, promoting more predictable but less creative output. Additionally, the Top-P value, which defines the cumulative probability distribution of tokens considered during generation, was set to 0.50. This combination supports coherent responses while still allowing for contextual variation. These parameters were selected in alignment with findings from Amin & Schuller (2024), who concluded that conservative configurations (temperature ≤ 0.3 and top-p ≤ 0.7) result in more stable and reliable model performance, whereas higher values tended to degrade output quality (ibid, p. 6).

To further scaffold scientific precision, we adapted the PICOT framework (Riva et al., 2012), traditionally used in clinical research, to structure question development in the context of biological phenomena. It consists of five components: Population (P), which defines the target group; Intervention (I), outlining the treatment being examined; Comparison (C), referring to the control condition or alternative intervention; Outcome (O), capturing the observable effects of the intervention; and Time (T), specifying the timeframe over which these effects are evaluated.

To support epistemic scaffolding the Socratic tutor was instructed to prompt learners to reframe and refine their research questions rather than offering pre-formulated answers. The uninstructed chatbot received the instruction to base answers on the PICOT framework but without an instruction to engage them in iterative questioning or reflective dialogue.

## 4. Methodology: Experimental Design and Mixed Methods Framework

To evaluate the Socratic AI Tutor, we conducted an experimental feasibility study in four seminars not taught by the authors, consisting of 65 pre-service teacher education biology students enrolled at a German university. Descriptive statistics on sample characteristics are summarised in

Table **2**. Furthermore, we are planning to extend the number of participants in the upcoming semester.

**Table 2**

*Baseline characteristics*

|  | Total (n=65) | Socratic Tutor (n=36) | Non-socratic Tutor (n=29) | p-value |
|---|---|---|---|---|
| **Female (mean(SD))** | 0.62 (0.49) | 0.61 (0.49) | 0.62 (0.49) | 0.938 |
| **Age (mean (SD))** | 23.45 (4.98) | 23.28 (5.19) | 23.66 (4.81) | 0.762 |
| **No. of semester (mean (SD))** | 4.72 (4.57) | 4.78 (5.39) | 4,66 (3.36) | 0.911 |
| Joint Orthogonality Test |  |  |  | 0.985 |

Note Pillai's trace MANOVA tested for joint differences in gender, age, and semesters between conditions.

The study employed a between-subjects design with two experimental conditions: one group interacted with the Socratic AI Tutor (ST), while the control group engaged with an uninstructed, general-purpose AI chatbot (nST). Two seminars were allocated randomly to each, Condition A and B.

The study comprised four sequential phases. First, all participants completed an initial test based on a validated item-battery[1] for research question formulation competency of students in biology (Mathesius et al., 2014) and a baseline task in which they independently formulated a research question in response to a biological phenomenon without AI support. All phenomena can be found in annex1. This within-subjects measure provided a performance anchor prior to treatment.

Second, participants were asked to interact with their assigned AI system to iteratively refine their initially stated research question. AI interactions were time-limited to 5 minutes to standardise exposure.

Third, students were asked to fill out a post-intervention survey based on the Unified Theory of Acceptance and Use of Technology (UTAUT2; Venkatesh, 2022) and the AI-chatbot perception scale (Stöhr et al., 2024) , adapted to the educational AI context (Annex 2). The instrument captured, among others, perceived usefulness, effort expectancy, social influence, and behavioural intention to reuse the system.

---

[1] Ko-WADiS (Kompetenzmodellierung und -erfassung zum Wissenschaftsverständnis über naturwissenschaftliche Arbeits- und Denkweisen bei Studierenden (Competence modelling and assessment of scientific literacy through scientific methods and modes of thinking in university students).

Fourth, two transfer tasks were included to assess the ability to apply learned skills in new contexts. One task involved near transfer, presenting a structurally similar scenario, while the second required far transfer, asking students to formulate a research question within a different biological domain (e.g. plant anatomy), without AI assistance. The distinction follows the definitions of near and far transfer as outlined by Barnett & Ceci (2002). The plant anatomy context was selected based on its alignment with the students' curriculum at that point in the semester.

At the time of writing, the analysis of students' research question quality has not yet been completed. Responses will be rated on a 5-point Likert scale by two trained experts working independently. Interrater reliability will be calculated using Fleiss' Kappa and reported in a subsequent manuscript. The present paper therefore focuses on the second strand of our mixed methods design: participants' perceptions of the chatbot's support across different domains.

To ensure that post-intervention effects were not attributable to baseline differences in scientific reasoning, we conducted an analysis of the pre-test performance across the two conditions. Total scores ranged from 0 to 3. A Welch's t-test revealed no statistically significant difference in pre-test performance between the Socratic ($M = 0.57$) and non-Socratic condition ($M = 0.53$), $t(115.07) = -0.29$, $p = .773$, 95% CI [$-0.34, 0.25$]. The effect size was negligible, $d = -0.05$, 95% CI [$-0.41, 0.30$]. An intraclass correlation coefficient (ICC) was also computed to assess variance explained by group membership The result, ICC = $-0.015$ confirmed that virtually none of the variance in pre-test scores could be attributed to group membership. Together, these results rule out pre-existing differences in factual knowledge or reasoning ability as plausible confounds.

# 5. Findings I: User Perception on Support for Critical, Independent, and Reflective Thinking

To explore participants' subjective experience of AI support, we asked students to rate the extent to which the chatbot fostered critical, independent, and reflective thinking. Across all three dimensions, students in the Socratic condition perceived significantly greater support than those using the uninstructed chatbot (summarised in Fig. 1 and Table 3). Regarding the item EKICQ2.SQ006 [To what extent do you agree with the following statement? "The answers of the AI chatbot stimulated reflection." (1: I completely agree; 5: I do not agree at all)][2] a refined multi-level analysis that accounts for participant-level variance through random intercepts, revealed a statistically significant difference in participants' ratings of how thought-provoking the chatbot responses were, with model yielding a significant group effect ($b = -0.66$, $SE = 0.22$, $t(62.48) = -2.97$, $p = .004$)) and a standardised coefficient of $\beta = -0.54$, 95% CI [$-0.94$, $-0.13$].

This suggests that students in the Socratic condition found the chatbot's responses more thought-provoking. Similarly, we found statistically significant difference in students' agreement with the statement "The chatbot fosters my critical thinking", the Socratic group again reported significantly higher perceived support ($b = -0.61$, $SE = 0.23$, $t(65.27) = -2.62$, $p = .011$), with a standardised effect of $\beta = -0.54$, 95% CI [$-0.94$, $-0.13$]. This suggests a moderate effect size consistent with the tutor's intended design. Supporting the trend, the linear mixed-effects model revealed a robust effect of condition, with participants in the Socratic group rating the chatbot as significantly more motivating for supporting independent thought ($b = -1.18$, $SE = 0.23$, $t(65.02) = -5.12$, $p = <.001$). The standardised coefficient was $\beta = -0.96$, 95% CI [$-1.33$, $-0.59$], indicating a large effect. This is in line with results from interviews conducted during previous tests as participant 00CA6 reported:

---

[2] Original question in German: EKICQ2.SQ006: In welchem Ausmaß stimmen Sie den folgenden Aussagen zu? Die Antworten des KI Chatbot haben zum Nachdenken angeregt. (1: Ich stimme voll und ganz zu; 5: Ich stimme überhaupt nicht zu)

It was striking that the chatbot basically never told me anything but only gave me food for thought so that I could find a solution myself. And it encouraged me to think the whole time, to reflect on myself, to think about my own sentences that I had given the chatbot and to think about them again myself. As a result, the chatbot was more of a guiding form, without giving much content or making many suggestions, but rather going into more detail and providing food for thought through many small questions. (translated[3])

**Fig. 1**

*Box-plot of Results Comparing Socratic Tutor and Non-Socratic Tutor Groups on Perceived AI Chatbot Support for Critical, Independent, and Reflective Thinking*

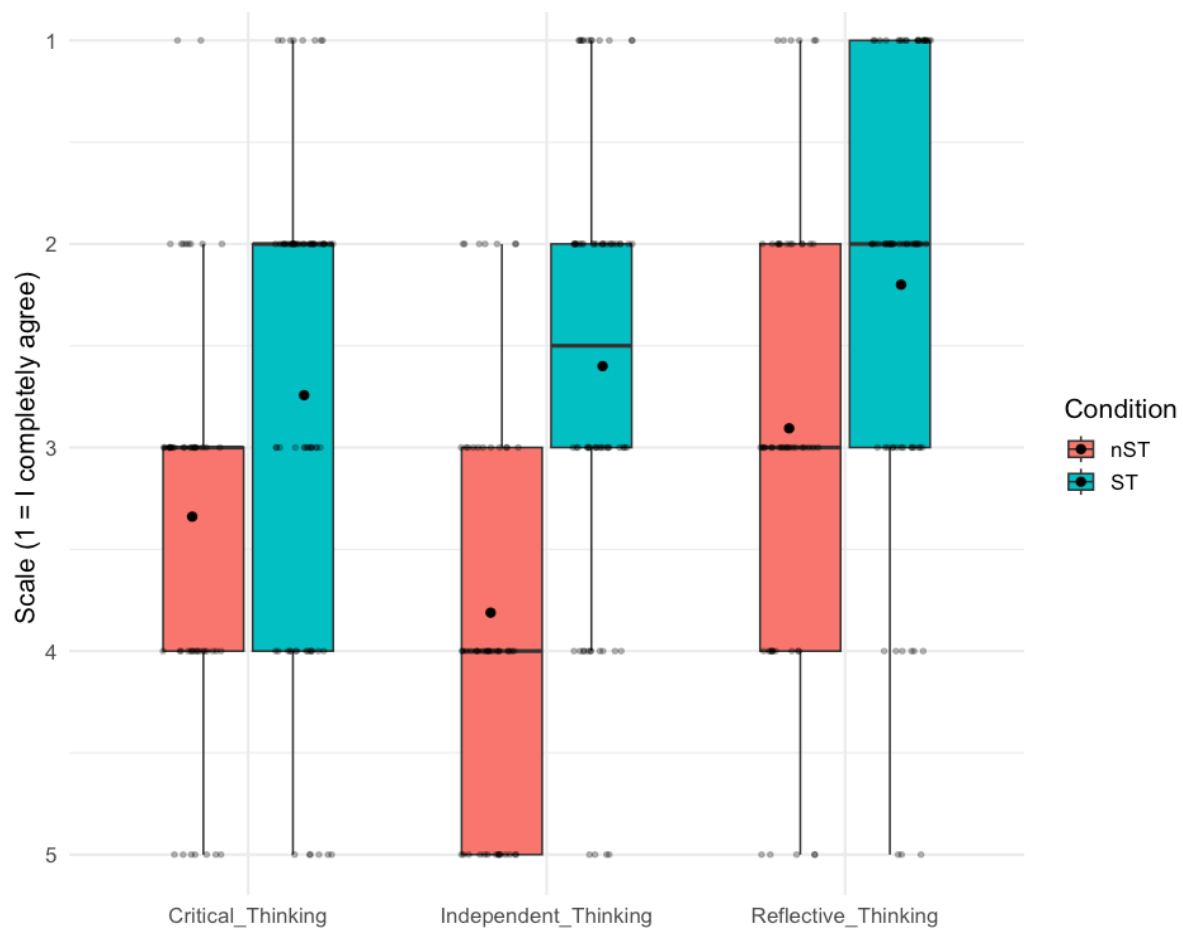

---

[3] Original German answer: Auffallend war, dass der Chatbot mir im Grunde nie irgendwas vorgesagt hat, sondern mir quasi nur Denkanstöße gegeben hat, damit ich es selber zu einer Lösung finde. Und er hat halt die ganze Zeit zum Denken angeregt, dass ich mich selbst reflektiere, meine eigenen Sätze, die ich dem Chatbot gegeben habe, nochmal durch den Kopf gehen lasse und mir selber nochmal Gedanken mache. Und dadurch war der Chatbot eher in einer leitenden Form, ohne jetzt groß inhaltlich was vorzugeben oder viele Vorschläge zu machen, sondern eher durch viele kleine Fragen nochmal näher ins Detail gehend und Denkanstöße liefernd.

**Table 3**

*Multi-level Analysis Results Comparing Socratic Tutor and Non-Socratic Tutor Groups on Perceived AI Chatbot Support for Critical, Independent, and Reflective Thinking*

| Item | Predictor | Estimate | Std. Error | df | t | ß | 95% CI (ß) |
|---|---|---|---|---|---|---|---|
| Critical Thinking | Intercept | 3.364 | 0.175 | 66.83 | 19.17 | 0.32 | [ 0.01, 0.63] |
| | Condition [ST]* | -0.612 | 0.234 | 65.27 | -2.62 | -0.54 | [-0.94, -0.13] |
| Independent Thinking | Intercept | 3.794 | 0.173 | 66.50 | 21.97 | 0.55 | [ 0.27, 0.83] |
| | Condition [ST]*** | -1.179 | 0.230 | 65.02 | -5.12 | -0.96 | [-1.33, -0.59] |
| Reflective Thinking | Intercept | 2.878 | 0.167 | 64.28 | 17.21 | 0.32 | [ 0.04, 0.61] |
| | Condition [ST]** | -0.661 | 0.222 | 62.48 | -2.97 | -0.57 | [-0.95, -0.19] |

*Note.* *p < .05, **p < .01, ***p < .001. Score ranges from 1 (I completely agree) to 5 (I do not agree at all), n=36 (ST) & 29 (nST). Questions: To what extent do you agree with the following statement? (1) The chatbot fosters my critical thinking. (2) The chatbot motivates me to think independently. (3) The answers of the AI chatbot stimulated reflection. Underlying values reflect raw item means; inferential results were obtained via multi-level modelling using R (R version 4.3.0, library(lme4), Linear mixed model fit by REML ['lmerMod'], random intercepts by participant, 65 students, 123 observations. Standardised coefficients calculated using Satterthwaite's method ['lmerModLmerTest']. Standardised coefficients (β) were computed using full z-standardisation of both outcome and predictor variables and refitting the model accordingly (method = "refit").

In addition to the pre-tests described in section 4, we examined whether observed differences could be attributed to the timing of the intervention rather than the experimental manipulation. We extended the linear mixed-effects models by introducing random intercepts for both the scheduled group session (t1–t3) and the actual individual session attended by each participant. This distinction accounts for cases in which participants could not attend their originally assigned session (e.g. due to illness) and instead completed their individual session (e.g. session 1) during a different group session (e.g. group session 2). The results of the individual session-based models are shown in Fig. 2. This allowed us to account for potential clustering effects related to when participants engaged with the chatbot. Across all three outcome variables session-level variance was estimated as zero or near-zero, and model fit remained unchanged. These results indicate that the timing of participation did not systematically influence perceived support for thinking, lending further robustness to the condition-level effects observed.

**Fig. 2**

*Box-plot of Results Comparing Socratic Tutor and Non-Socratic Tutor Groups on Perceived AI Chatbot Support for Critical, Independent, and Reflective Thinking across participants' sessions.*

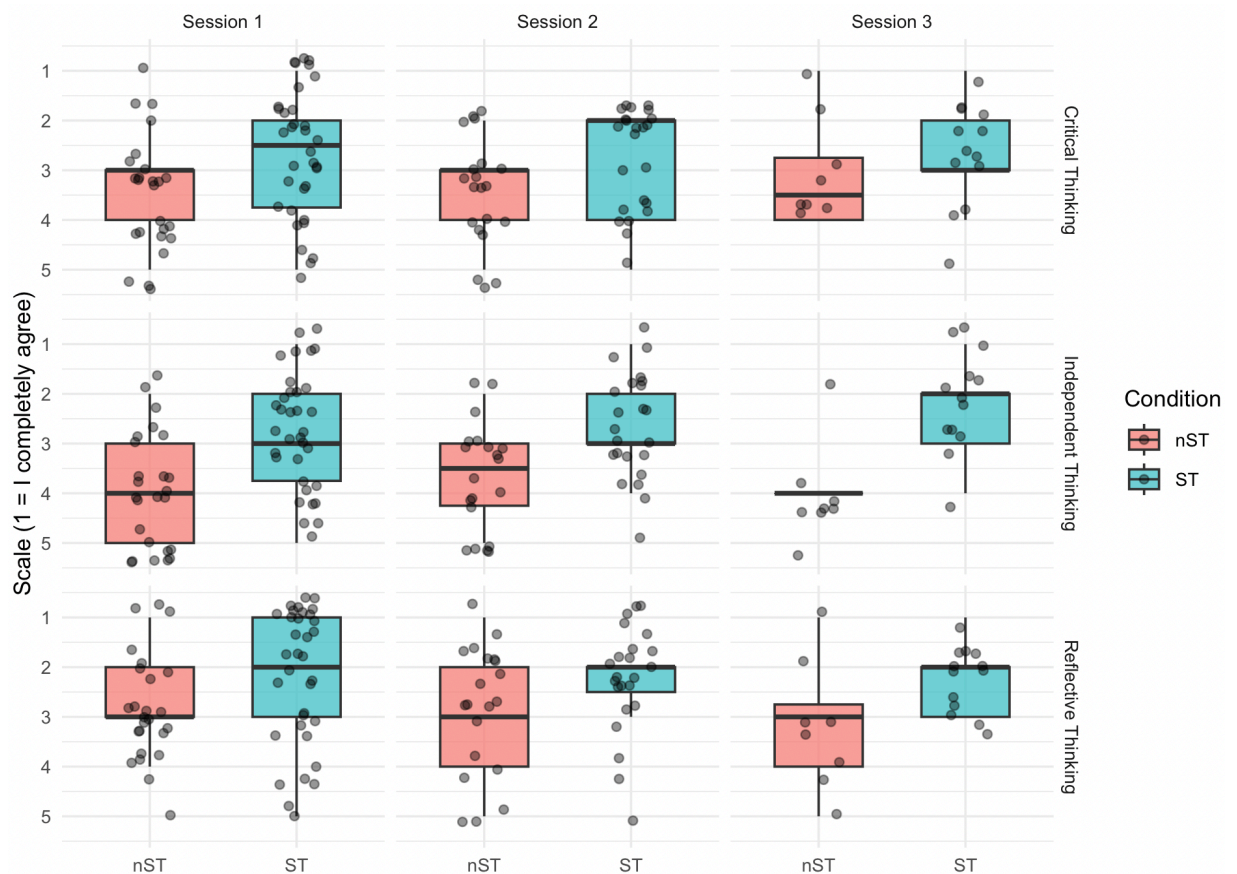

*Note.* Score ranges from 1 (I completely agree) to 5 (I do not agree at all), n=36 (ST) & 29 (nST). Questions: To what extent do you agree with the following statement? (1) The chatbot fosters my critical thinking. (2) The chatbot

*motivates me to think independently* (3). The answers of the AI chatbot stimulated reflection. Underlying values reflect raw item means; inferential results were obtained via multi-level modelling using R (R version 4.3.0, library(lme4), Linear mixed model fit by REML ['lmerMod'], random intercepts by participant), 65 students, 123 observations.

Nevertheless, a closer inspection of post-intervention responses reveals a more nuanced pattern. While the overall average effect of the Socratic chatbot on perceived thinking support was robust, students perceive their experience differently. A small number of students in the Socratic condition strongly disagreed with statements such as "The chatbot motivated me to think independently." While the tutor generally promoted deeper cognitive engagement, it was not equally effective for all learners. Such outliers may reflect individual differences in prior beliefs, engagement, or familiarity with AI tools, and highlight the importance of personalisation and adaptability in the design of AI-mediated learning experiences.

## 6. Consequences: Toward Multi-Agent Learning Architectures

While the Socratic tutor demonstrated its effectiveness in fostering epistemic agency, particularly in the formulation of research questions, it also reveals the inherently limiting characteristics of special-purpose AI. Contemporary AI-enhanced learning systems often operate as standalone specialist agents (Bašić et al., 2023; Du & Daniel, 2024; Liu et al., 2025; Menekse et al., 2024; Morris et al., 2024; see also Moldoveanu (2025, p. 153ff for a list of specific roles in a Tertiary Education Environment). The Socratic AI tutor evaluated in this study exemplifies such a system: effective in its domain of structured research question refinement but disconnected from other crucial dimensions of the learning experience such as emotional support or learning trajectory guidance. This monolithic deployment model, i.e. operating in isolation rather than in a multi-agent system (MAS), lacks the flexibility and responsiveness required to support the full spectrum of learner needs.

Notably, the term *multi-agent systems* can be understood in two distinct yet complementary ways. On the one hand, it refers to multiple AI agents collaboratively contributing to a single, well-defined pedagogical function. This is can be exemplified by J. Zhang et al. (2025), whose framework integrates Socratic guidance within an ensemble of seven agents, each with distinct functionalities, in their **M**ulti-**A**gent framework Inco**r**po**R**ating **S**ocratic guidance (MARS). Similarly, IntelliChain framework by Qi et al. (2025) integrates several LLMs within a single domain.

On the other hand, the term also encompasses the idea of a broader ecosystems of specialised agents, each designed for different educational tasks, operating in concert across institutional contexts as described above. For instance, the work of Agrawal & Nargund (2025) on a framework for optimal agent selection points in this direction. This interpretation reflects a systemic vision of AI integration, where orchestration across diverse agent roles enables adaptive, modular support for heterogeneous learner needs.

To reduce conceptual ambiguity, we propose distinguishing between *coordinated* and *orchestrated* multi-agent systems. Here, coordinated MAS refer to tightly integrated agent ensembles that operate within a single pedagogical domain or task. For instance, emotional support, conceptual scaffolding, or critical feedback. In such systems, agents may assume complementary subroles (e.g., monitoring, prompting, or regulating) to enhance a shared function. In contrast, orchestrated MAS describe more expansive configurations in which specialised agents span across domains, operating in loosely coupled but pedagogically synchronised ways. These orchestrated systems constitute a modular AI ecosystem, wherein agents interact not merely for task optimisation but to sustain diverse learning trajectories and holistic student development.

Building on this insight, we argue that the next step in AI-enhanced learning is not necessarily further optimisation of singular agents, but the design of orchestrated MAS. On a broader level orchestrated MAS could enable scalable yet personalised constellations of support, allowing students to access responsive guidance outside formal lecture, seminar or office-hour contexts. Services provided to students by specialists with limited availability might be transferred to a specialised AI agent. For example for psychological support (Broglia et al., 2018; Hill & Key, 2019) or career advice (Brown et al., 2019; Dey & Cruzvergara, 2014; Khurumova & Pinto, 2024), consultation on academic writing, proof reading of CVs or interview coaching, as exemplified by the services at leading universities such as the LSE (London School of Economics and Political Science, 2025) or UCL (University College London, 2025).

To visualise this, we propose an adaptation of offer-and-use models (Helmke, 2009; Lipowsky, 2015), originally stemming from the contexts of school and teacher professional development, respectively, for AI-supported learning (Fig. 3). In our adaptation, orchestrated AI agents function as differentiated offers. The learner, guided by human and non-human facilitators, selects and appropriates these offers based on instructional context and personal learning needs.

**Fig. 3**

*Offer and Use Model of Higher Education in Times of Orchestrated Multi-Agent-Systems*

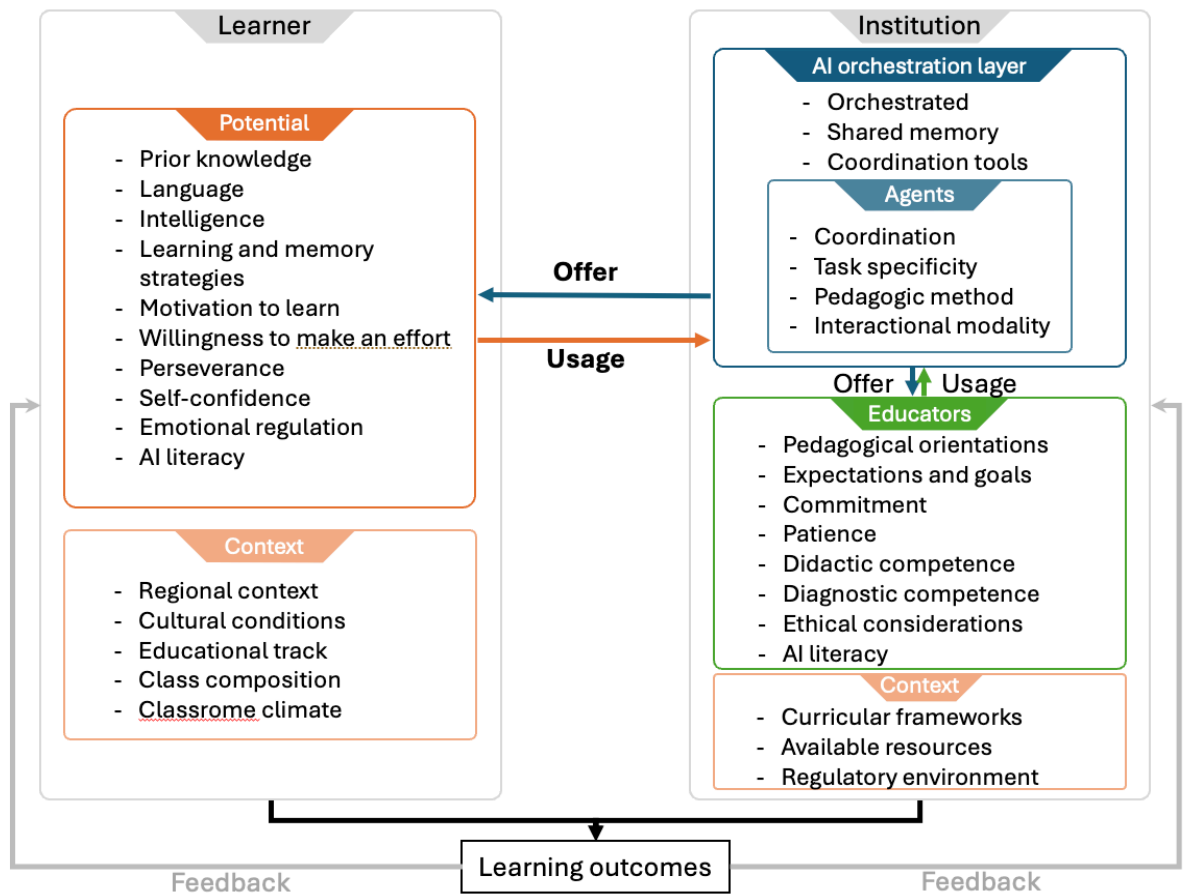

*Note.* Adapted from offer-and-use models by Helmke (2009) & Lipowsky (2015).

This modular logic reflects a long-standing principle in educational design: the division of pedagogical labour (McMillan et al., 2016). Just as human teachers, mentors, and counsellors occupy distinct roles, so too can AI agents contribute specialised expertise within orchestrated constellations. The shift from standalone agents to orchestrated ecosystems allows for greater pedagogical granularity and adaptability.

To illustrate how modular orchestration might be implemented in practice, consider a hypothetical system for thesis supervision:

- A Socratic tutor agent facilitates the initial formulation of research questions and ensures logical coherence.
- A critical feedback agent delivers iterative evaluations on argument structure and the integration of evidence.
- An affective support agent tracks motivational and behavioural engagement patterns, such as writing frequency, timing, or progress indicators, and offers reflective prompts or evidence-based coping strategies.

All agents operate within a shared learner model and log interactions to a transparent orchestration dashboard, accessible to both student and supervisor. Utilising cross-agent memory persistence, as explored in MARS (J. Zhang et al., 2025) might allow for maintaining coherence without sacrificing modularity.

However, this shift also raises critical questions: When should instructional responsibility be delegated to AI agents? How must curricula evolve to reflect an emerging paradigm? How should the organisational logic of higher education institutions be reconfigured in light of AI orchestration?

## 7. System-level Implications part 1: Institutions. Faculty roles, Curricula and Assessments, institutional logics.

Answering these questions necessitates a re-evaluation of the system-level implications for higher education institutions against the background of multi-agent learning architectures. Hence, this chapter addresses the emerging contours of this transformation across three critical dimensions: the division of pedagogical labour, curricular implications and institutional infrastructure. Together, these areas capture the broader shifts required as AI systems transition from isolated tools to integrated components of HEI. Rather than treating orchestration as a purely technical challenge, we examine how it reconfigures the roles of educators, reshapes what students learn, and demands new forms of institutional coordination and investment.

### 7.1. Rethinking Division of Pedagogical Labour: Educators as Users of the AI Orchestration Layer

Historically, there has been a low level of division of pedagogical labour at HEIs, relying mainly, especially in central Europe: "The German system was monocratic; a single professor was responsible for his whole subject, no one shared in his authority in the distribution of teaching (…)" (Shils & Roberts, 2006). However, industrial-era models of education built on one-teacher–many-students configurations, i.e. the assumption of a single instructor delivering content to many students, appear ill-suited to the emerging ecology of multi-agent systems. In these new constellations, faculty members are no longer the exclusive source of explanation, critique, or support. Instead, they are becoming *orchestrators* of agent-mediated learning processes.

Building on the previously mentioned example, delegating the initial formulation of research questions to a Socratic tutor, not only conserves instructional resources but could also serve as a productive precursor to more focused supervisory dialogue. Similarly, other specialised agents, such as critical feedback companions or style advisors could be integrated in the learning process by the educator to allow students to pre-structure their work to a level of formal and conceptual adequacy before interpersonal teaching. This does not replace supervision but recalibrates it, shifting from basic troubleshooting or structural correction toward higher-level epistemic engagement.

Such a redistribution also holds promise in terms of equity and scalability. As generative AI increasingly exhibits cognitive performances that rival or exceed those of average learners and even staff (Carolus et al., 2025; Kestin et al., 2024; Luo et al., 2025; Salvi et al., 2025), the question is no longer whether these tools can "help" students, but rather how educators should reposition themselves within this altered ecology. Here, orchestration becomes a core professional competency as educators curate, sequence, and coordinate the availability and interplay of diverse agents in alignment with learning objectives. It also entails a rethinking of temporalities as learners can move beyond episodic contact hours to continuous, adaptive learning flows mediated by agent interaction.

This orchestration should involve diagnostic discernment (which agent for which learner, at which moment), ethical oversight (e.g. guarding against bias (Ferrer et al., 2021; Idowu et al., 2024; Ouyang et al., 2023), dependency (Buçinca et al., 2021; Chen, 2025), or de-skilling (Fan et al., 2025; Shukla et al., 2025; Vallor, 2015), and the cultivation of agency among students. In this regard, Southworth et al. (2023), drawing on Zimmermann (2018), emphasise that "It is particularly the responsibility of educators – who are generally reflective practitioners – to understand the ramifications of implementing AI in the educational system and to take active steps (…)" (p. 3).

Moreover, to guide this orchestration in a pedagogically responsible manner, educators must decide when and how to delegate instructional responsibility to AI agents. Recent research suggests two main preconditions for such decisions:

First, task clarity. AI agents excel when operating within well-defined, rule-bound domains, such as rhetorical structuring or reference formatting. Delegation is most effective when the instructional goal can be formalised and scaffolded through algorithmically tractable patterns (Butterfuss et al., 2022; Guggenberger et al., 2023). For example reviews identified studies that proved the usage of AI agents for feedback on writing as useful (Manhiça et al., 2022; Ouyang et al., 2022). Tasks that demand ethical reasoning by contrast, still call for human leadership (Bond et al., 2024) in line with discussion on ethics in AI in general (Holmes et al., 2022).

Second, the alignment between educator intent and the provided agents' capability must be explicit. Drawing on principal–agent theory (Hadfield-Menell, 2021), delegation is most effective when educators remain aware of the agent's strengths, limitations, and operational assumptions. This becomes particularly salient when the principal–agent dynamic begins to invert as agents not only receive instructional tasks from educators, but also start shaping the student's own task structuring or delegation decisions (Guggenberger et al., 2023).

As agents begin to assume more autonomous roles in shaping student inquiry and epistemic routines, educators must be equipped not only to delegate effectively, but also to interrogate, calibrate, and re-align agentic contributions within evolving pedagogical frameworks. This calls for targeted training programmes that move beyond basic AI literacy, fostering critical competencies in agent orchestration, prompt engineering, and diagnostic oversight.

## 7.2. Curricular Alignment and New Assessment Logics for Multi-Agent Learning Systems

The emergence of generative AI as a general-purpose multi-agent infrastructure not only effects educators but also students and thus compels a fundamental rethinking of curriculum design in higher education. We argue that they must evolve to support a new epistemic paradigm. One where learners actively engage with, interrogate, and co-construct knowledge in collaboration with AI agents. This demands a pivot from "learning to know" to "learning to question", a shift that reorients educational purpose from reproduction to inquiry.

As Chen (2025) argues, generative AI now functions as epistemic infrastructure rather than merely a tool, shaping how learners access, interpret, and synthesise information. In this context, curriculum must embed competencies that equip students to critically engage with AI-generated content: to prompt effectively, assess relevance and accuracy, and identify embedded assumptions or biases. This entails developing not only technical fluency in interacting with AI, but also an awareness of how AI systems frame knowledge, what they omit, and when their outputs warrant further interrogation.

To foster such competencies, curricular architectures must shift from linear content delivery models to iterative, process-oriented designs. Existing models of progressive inquiry (Hmelo-Silver et al., 2007; Shulman, 1981) offer a promising foundation. These models emphasise cycles of hypothesis generation, explanation, and critique which map well onto human-AI co-inquiry. For example, instead of writing essays in isolation, students might iteratively refine arguments in dialogue with an AI tutor that challenges inconsistencies or suggests counter-positions, as demonstrated by the work of Wambsganss et al. (2021) on an adaptive dialog-based learning system for argumentation skill called "ArgueTutor".
Educators, in turn, assess the quality of this dialogic engagement, not just the final output.

This also requires a reconfiguration of assessment logics. Traditional evaluation methods emphasising static products are, in the author's opinion, ill-suited to agent-mediated learning processes as generative AI systems contribute substantively to the production of academic work and determining the extent of student authorship becomes progressively challenging. Indeed, recent empirical analyses have demonstrated the limited reliability of current detection tools, highlighting their susceptibility to both false positives and negatives, and their vulnerability to simple circumvention strategies (Weber-Wulff et al., 2023).

Instead, process-tracing assessments that capture how students interact with AI tools become essential. Portfolios of AI dialogues, annotated with student reflections on decision-making and critical turning points, can offer evidence of epistemic growth. For example, hybrid models such as the FACT (Fundamental, Applied, Conceptual, critical Thinking) framework for assessment (Elshall & Badir, 2025), reflecting the levels in Bloom's taxonomy (Bloom, 1956; Krathwohl, 2002), propose combining AI-supported tasks with AI-free baselines to safeguard foundational knowledge while recognising advanced applications and judgment.

In regard to curriculum development and assessment, the "AI Across the Curriculum" initiative reported on by Southworth et al. (2023) at the University of Florida (UF) offers a glimpse on how universities can systemically embed AI literacy across diverse academic programmes. Rather than isolating AI instruction within technical fields, the framework integrates AI competencies as cross-cutting curricular elements tied to disciplinary practices (Fig. 4).

**Fig. 4**
*AI Faculty Hiring Initiative*

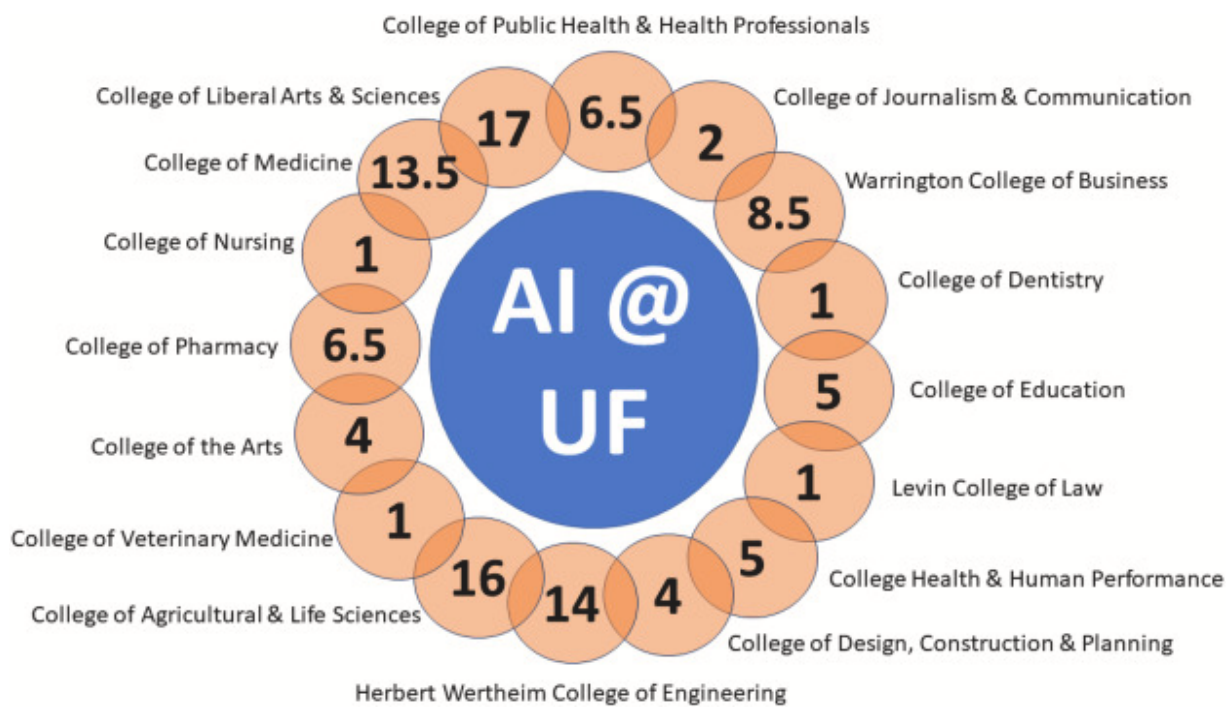

*Note.* Reprinted with permission from (Southworth et al., 2023, p.5) Original notes: AI Faculty Hiring Initiative – the college locations of the 106 faculty hired as part of the AI faculty initiative 2020–22, an ongoing process, with another 2 faculty in the Libraries and Florida Museum, and more hires currently underway. [Note 0.5 faculty indicates faculty in joint appointments across different colleges.].

Its structure rests on several pillars that reflect a holistic AI-approach. First, AI Medallion Programs incentivise undergraduate students to engage with AI in meaningful, interdisciplinary ways. These programs recognise curricular and extracurricular accomplishments in AI-related learning and signal student readiness to future employers (Southworth et al., 2023, p.8).

Second, the institution actively fosters partnerships with external stakeholders, including industry and public sector organisations, to embed real-world AI challenges into teaching, co-develop curricular offerings, and enhance career development pathways.

Third, the university is invested in curriculum development, coordinated through a dedicated AI academic initiative centre (AI$^2$), to ensure the integration of AI literacy across all faculties. This includes not only the design of new courses, but also the review and adaptation of existing offerings according to a refined AI literacy framework.

Fourth, the UF AI Literacy Model provides a coherent scaffold for student learning outcomes in regard to five categories: Enabling AI, Know & Understand AI, Use & Apply AI, Evaluate and Create AI as well as AI Ethics (ibid, p.7).

Fifth, a deliberate and well-resourced investment has been made into AI infrastructure, most prominently through the hey investments into AI-infrastructure. This addresses the proverbial chicken-and-egg problem where the availability of AI infrastructure must be matched by its meaningful pedagogical use to generate demand, which in turn justifies further investment.

The latter leads directly into the next section on institutional logics, which addresses how infrastructure, policy, and strategic governance must change against the background of multi-agent learning systems in higher education.

### 7.3. Institutional Logics

To provide the necessary infrastructure that would allow an AI orchestration layer readily usable by staff and students alike requires substantial upfront investment in technical infrastructure, such as high-performance computing clusters. Table 4 illustrates the associated costs, drawing on the example of the University of Florida, where the HiPerGator supercomputer underpins the comprehensive institutional strategy for AI integration.

**Table 4**

*Costs of University of Florida's HiPerGator supercomputer for AI from 2013 to 2025*

| Component | Cost (USD) |
|---|---|
| HiPerGator 1.0 hardware | $3.4 million |
| Data center facility (HiPerGator 1.0) | $15 million |

| Component | Cost (USD) |
|---|---|
| HiPerGator 2.0 upgrade/tuning | $8 million |
| HiPerGator 3.0 (CPU + storage)* | Part of AI initiative (see note) |
| HiPerGator AI (DGX SuperPOD) | $70 million |
| **Total through 2021 (incl. facility)** | **$96.4 million** |
| HiperGator 4.0 (planned) | $24 million |
| **Total including planned 2025 upgrade** | **$120.4 million** |

*Note.* The $70 million HiPerGator AI package included the 3.0 CPU/storage upgrade alongside the AI GPU expansion. Since UF did not break them out separately, we've allocated the full $70 million under the AI initiative. Nominal values are reported; the analysis does not account for changes in the purchasing power of money over time. Information has been collected from publicly available resources (Bergeron-Oakes, 2015; Martin, 2020; The Florida Senate, 2025)

Increasingly, such investments are moving beyond the scope of individual institutions, prompting a shift toward collaborative funding models and shared infrastructure arrangements to meet the scale and complexity demands of next-generation AI in academia. For example, the EuroHPC Joint Undertaking, a pan-European public–private partnership, has committed €7 billion (2021–2027) (European Commission, 2025), including national co-financing, to build shared supercomputing infrastructure accessible across institutional and national boundaries. This signals a broader recognition that the cost and governance demand of AI infrastructure exceed the capacity of many institutions acting in isolation. Against the background of more advanced multi-agent architectures, we expect this trend to continue.

For HEIs, such infrastructural consolidation and AI capability growth raise urgent ethical considerations. As Multi-Agent-Systems become embedded, the question is no longer whether ethics should be considered, but how ethics can be operationalised at design level. This calls for *Ethics by Design for AI* (Brey & Dainow, 2024).

In educational settings, Ethics by Design for AI entails aligning system functionality with ethical considerations such as fairness and transparency. Brey & Dainow (ibid) argue that ethical engagement cannot be external to system development. Rather, design decisions themselves structure what actions are possible, permissible, or prohibited. In this light, ethical learning systems should make their operations traceable, their decision-making interpretable, and their interventions contestable.

The functionalist perspective described by Dodig-Crnkovic et al. (2025) provides a useful approach for orchestrated AI systems. Within multi-agent environments, individual agents may carry limited or no moral accountability, yet their joint operations impact learning trajectories and educational fairness. Hence, ensuring "good behaviour in the future" (ibid, p. 3) becomes paramount. By distributing ethical safeguards, such as escalation protocols, oversight mechanisms, and role-specific accountability across agents and institutional processes, higher education can mitigate responsibility gaps while preserving system-level coherence. Importantly the "(…) delegation of responsibility to these systems does not absolve humans of their role. Instead, it requires a new framework where responsibility is dynamically shared. Humans are responsible for designing, overseeing, and updating the ethical features programmed in these systems, while AI is responsible for the ethical execution of tasks." (ibid, p. 4).

Last but not least, regulatory developments (Siegmann & Anderljung, 2022) indicate that the governance of AI in education is also becoming a matter of legal obligation. As institutions move toward more systematic integration of multi-agent systems, compliance with evolving legal standards will become an essential dimension of responsible implementation.

The European Union's AI Act represents a notable example of a structured regulatory approach. It categorises AI systems used in educational contexts, particularly those involved in evaluation of learning outcomes, monitoring or behavioural assessment, as high-risk applications. This classification triggers specific requirements, including requirements such as provisions risk mitigation, record-keeping, and human oversight (Regulation (EU) 2024/1689 of the European Parliament and of the Council, 2024). While the regulation is not limited to education, it underscores the view that learning environments warrant particular safeguards.

A parallel effort is visible in Australia, where the Tertiary Education Quality and Standards Agency (TEQSA) has issued national guidelines for managing generative AI in higher education. Developed by the TEQSA Australian Academic Integrity Network (AAIN) Generative AI Working Group, these guidelines emphasise both the opportunities and challenges posed by generative AI tools such as ChatGPT (AAIN, 2023). Among the key issues identified are threats to academic integrity, inequities in student access, and the need for sustainable, discipline-sensitive assessment practices. In response, TEQSA has mandated that all higher education institutions submit credible action plans outlining how they intend to ethically integrate generative AI technologies, with a particular emphasis on preserving the integrity of qualifications (Tran et al., 2025)

Together, these international developments reflect a shared regulatory trajectory in which educational systems move beyond reactive measures toward more proactive governance of generative AI. This includes not only regulatory compliance but also the institutional alignment of pedagogical practices and organisational structures to ensure the responsible and educationally sound integration of AI technologies.

## 8. System-level Implications Part 2: Students. Competencies and (Soft-)Skills

It is likely, that the integration of orchestrated multi-agent systems (MAS) into higher education introduces not only new technological infrastructure needs and faculty roles but also reconfigures the expectations of the very competencies that students are to develop. While traditional paradigms of higher education have centred on individual cognition, subject-specific mastery, and the cultivation of disciplinary knowledge [SOURCES!], multi-agent ecosystems shift the emphasis toward co-agency, dialogic interaction (from learning to know to learning to question), and adaptive learning trajectories as AI gives support for epistemic co-construction on-par with human interactions (Haase & Hanel, 2023). This section outlines the anticipated impact of multi-agent systems on student competencies, building on findings from the experiment and recent literature from the field of AI in education.

A key finding of the experiment was the statistically significant increase in students' perceived development of critical, independent, and reflective thinking when supported by the Socratic AI tutor. These outcomes suggest that dialogic AI systems can scaffold deeper cognitive engagement, especially when designed to elicit metacognitive reflection rather than deliver factual responses. Other research points in the same direction:

Huang et al. (2024), for instance, conducted a small study in the context of student led project development for health promotion. Their findings suggest, that integrating a generative AI assistant into student design projects can significantly boost creative and design thinking outcomes, as the AI's suggestions prompts students to consider more novel solutions. Similarly, Yao et al. (2025) found significant improvements regarding teamwork abilities and problem-solving skills by introducing generative AI in the 12 weeks course.

Students themselves recognize these benefits as several surveys find, that university students noted that ChatGPT and similar tools provide helpful assistance in areas such as idea generation and writing, supporting their creative tasks e.g. HongKong, Chan & Hu, 2023).

In multi-agent architectures, these effects can be further amplified as students may engage with diverse agents fulfilling specialised roles. This multiplicity introduces learners to a distributed cognitive ecology, wherein they must negotiate perspectives, reconcile conflicting suggestions, and take responsibility for knowledge synthesis to a higher degree than in currently. The result is a higher degree of flexibility but also the need for AI literacy: the ability to move between different epistemic positions, critique AI-generated content, and integrate contributions across agents.

### 8.1. A Paradox? Renewed Importance of Traditional Competencies and 21st Century Competencies

On the one hand, despite the shift toward co-agency, dialogic interaction, and adaptive learning trajectories and perhaps paradoxically, the rise of orchestrated MAS may lead to a renewed importance of traditional academic competencies. As students interact with agents that simulate expertise or offer novel perspectives, they must also possess the domain-specific grounding necessary to evaluate the plausibility, coherence, and relevance of AI-generated content, as AIs can generate misleading or false information (Y. Zhang et al., 2023).

Far from rendering content knowledge obsolete, these systems amplify the need for critical disciplinary literacy, comprising for instance the capacity to verify claims, cross-reference interpretations, and spot fallacies or concept drift (Webb et al., 2017). The concern, that AI might be a barrier to education is also partially reflected by the results of our post-intervention survey (summarised in Fig. 5 and Table 5), as a significant part of the surveyed students expressed concerns. While most differences between conditions were not statistically significant, the effect for future learning risk approached conventional thresholds ($b = 0.47$, $p = .062$, $\beta = 0.41$), suggesting a potential trend that may warrant further investigation. This result also suggests that scaffolded, dialogic engagement may mitigate perceptions that AI inherently undermines educational values, an outcome consistent with the idea that epistemic partnership, can foster more constructive attitudes toward generative tools in academic settings (Huang, 2024; M. Liu et al., 2024; Wambsganss et al., 2025).

**Fig. 5**

*Box-plot of Results Comparing Socratic Tutor and Non-Socratic Tutor Groups on Perceived Concerns About AI Chatbot Use*

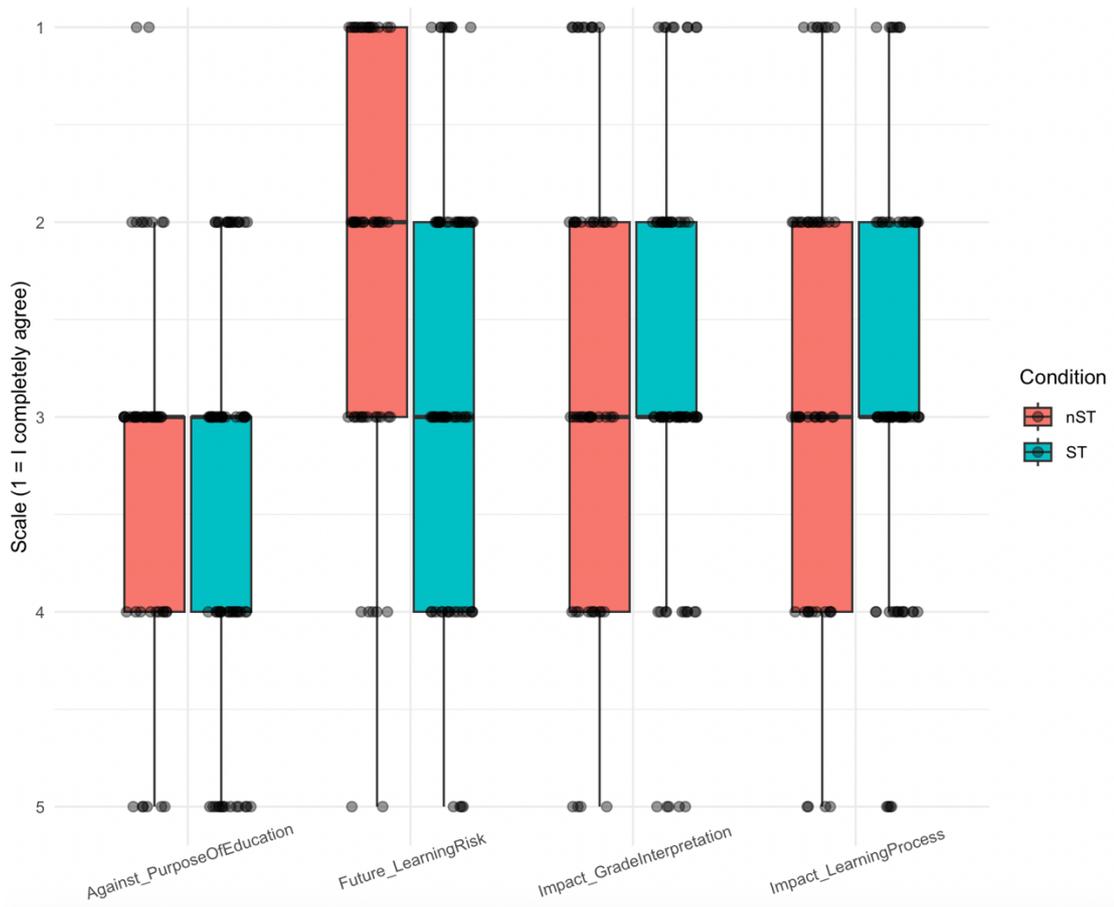

*Note.* Score ranges from 1 (I completely agree) to 5 (I do not agree at all), n=36 (ST) & 29 (nST). Questions: To what extent do you agree with the following statement? (1) The use of AI chatbots contradicts the purpose of education. (2) I am concerned about how AI chatbots will affect student learning in the future. (3) I am concerned about how the use of the chatbot could negatively affect my learning process. (4) I am worried that my own performance becomes less meaningful due to the use of AI. The answers of the AI chatbot stimulated reflection. Underlying values reflect raw item means; inferential results were obtained via multi-level modelling using R (R version 4.3.0, library(lme4), Linear mixed model fit by REML ['lmerMod'], random intercepts by participant), 65 students, 490 observations.

**Table 5**

*Multi-Level Analysis Results Comparing Socratic Tutor and Non-Socratic Tutor Groups on Perceived Risks of AI Chatbot Use*

| Item | Predictor | Estimate | Std. Error | df | t | ß | 95% CI (ß) |
|------|-----------|----------|------------|------|-------|-------|----------------|
|      | Intercept | 3.175    | 0.168      | 65.56| 18.94 | -0.15 | [-0.47, 0.18]  |

| Item | Predictor | Estimate | Std. Error | df | t | ß | 95% CI (ß) |
| --- | --- | --- | --- | --- | --- | --- | --- |
| Against Purpose of Education | Condition [ST] | 0.259 | 0.223 | 63.72 | 1.16 | 0.25 | [-0.18, 0.69] |
| Future Learning Risk | Intercept | 2.257 | 0.184 | 63.05 | 12.27 | -0.25 | [-0.57, 0.07] |
| | Condition [ST] | 0.467 | 0.245 | 61.24 | 1.9 | 0.41 | [-0.02, 0.84] |
| Learning Process | Intercept | 2.723 | 0.177 | 65.29 | 15.36 | -0.08 | [-0.41, 0.24] |
| | Condition [ST] | 0.185 | 0.237 | 63.81 | 0.78 | 0.17 | [-0.26, 0.60] |
| Performance Interpretation | Intercept | 2.698 | 0.189 | 64.07 | 14.27 | -0.04 | [-0.38, 0.29] |
| | Condition [ST] | 0.148 | 0.253 | 62.78 | 0.59 | 0.13 | [-0.31, 0.57] |

*Note.* Score ranges from 1 (I completely agree) to 5 (I do not agree at all), n=36 (ST) & 29 (nST). Questions: To what extent do you agree with the following statement? (1) The use of AI chatbots contradicts the purpose of education. (2) I am concerned about how AI chatbots will affect student learning in the future. (3) I am concerned about how the use of the chatbot could negatively affect my learning process. (4) I am worried that my own performance becomes less meaningful due to the use of AI. Underlying values reflect raw item means; inferential results were obtained via multi-level modelling using R (R version 4.3.0, library(lme4), Linear mixed model fit by REML ['lmerMod'], random intercepts by participant, 65 students, 122 observations. Standardised coefficients calculated using Satterthwaite's method ['lmerModLmerTest'].

On the other hand, as universities have long functioned not merely as venues for knowledge transmission but as spaces for cultivating human connection and community building (Brouwer et al., 2016; Pittman & Richmond, 2008), the shift toward AI-mediated learning necessitates a conscious reaffirmation 21$^{st}$ century competences such as collaboration, communication and creativity as summarised by Voogt & Roblin (2012). Dialogic engagement, collaborative learning, and informal peer interaction, hallmarks of the university experience, must be preserved and reimagined within MAS ecologies. Interacting with AI agents should not displace, but rather scaffold and extend, the meaningful dialogue between learners.

In sum, we expect the introduction of orchestrated MAS in higher education to not only augment instructional capacity but reconfigure what it means to be a competent learner. Future-ready students must be able to collaborate with, reflect on, and critique AI systems while cultivating their own creativity and social reasoning. Multi-agent environments therefore provide both a challenge and an opportunity: they invite new forms of interaction that stretch traditional notions of competence but require equally new pedagogical commitments to ensure these interactions remain equitable and educationally meaningful.

## 9. Cost-Effectiveness and Scalability Considerations

The average cost per student interaction with the Socratic AI chatbot during the experiment was approximately $0.0057 for a five-minute session, resulting in a total expenditure of $0.25 across all participants. These figures reflect token-based pricing and are subject to variation depending on the length and complexity of individual prompts and responses. Given equivalent token-based pricing, the Socratic AI tutor proved more cost-effective than the non-Socratic variant. In light of recent evidence that online tutoring can generate significant learning gains (Carlana & La Ferrara, 2024), our findings highlight the promise of dialogic, AI-mediated scaffolding. Socratic AI tutors may offer a scalable and cost-efficient alternative that aligns with pedagogical goals traditionally associated with human tutoring.

Additionally, for comparative purposes, we estimated the cost of an equivalent human-led tutorial interaction using a conservative calculation based on the hourly rate of a research associate in the German higher education context. Excluding ancillary employment costs (e.g. social security contributions), and assuming a gross salary-based hourly rate (pay scale TV-L 2025, E13,3), a five-minute exchange per student would amount to approximately €2.72. Multiplied across 44 students, this results in an estimated total of €119.68[4] for a given HEI replicating this experiment with humans.

---

[4] 
$$Cost = \left(\frac{Gross\ salary}{Hours\ per\ month} * \frac{1}{60}\right) * Number\ of\ minutes * Number\ of\ students$$

This comparison illustrates the substantial cost differential between AI-supported and human-mediated forms of individualised support, offering a preliminary basis for evaluating cost-effectiveness in scaling orchestrated MAS. However, while these figures demonstrate the comparative affordability of AI-based support, they are based on a specific implementation using proprietary software under commercial licensing. As institutions explore more complex forms of support, such as orchestrated multi-agent systems (MAS), cost projections must account for potential increases related to software subscriptions, licensing fees, and API usage caps. These concerns raise important questions about long-term scalability and institutional dependence on commercial vendors.

From an HEI standpoint, the true value may instead lie in the collaborative development of robust, open-source agents and orchestration frameworks, i.e. resources that can be adapted to institutional needs without recurring licensing constraints (Geiecke & Jaravel, 2024; Ghioni et al., 2024; Han et al., 2025; Langenkamp & Yue, 2022). In connection with a collaborative approach to infrastructure, such as the one outlined in section 7.3 referencing the EuroHPC Joint Undertaking, providing access to orchestrated multi-agent systems (MAS) may become feasible even for institutions with limited technical capacity. Unlike vertically integrated on-premises models (e.g. the University of Florida's HiPerGator), which require comprehensive in-house development and substantial financial commitment, shared infrastructure models could allow higher education institutions (HEIs) to pool resources, distribute development costs, and build upon common open standards.

Finally, though initial investments in faculty training and system familiarisation with orchestrated MAS might appear steep, these could be substantially reduced through modular self-paced learning formats. Such training not only limits operational costs but also ensures pedagogical alignment and sustained institutional capacity.

## 10. Conclusion: A New Learning Infrastructure for Higher Education

This paper set out to move beyond dominant automation narratives by reframing generative AI not as a tool for cognitive offloading, but as a dialogic infrastructure capable of augmenting epistemic agency. Drawing on constructivist theory and the Socratic method, we presented an AI chatbot prototype designed to support students in the formulation of research questions through structured questioning. Empirical findings from our controlled experiment underscore that such dialogic AI systems can meaningfully stimulate reflective, critical and independent thinking.

While this experiment serves as feasibility assessment and provide evidence in favour of dialog infrastructure, the significance of this intervention lies not in the standalone performance of the Socratic Tutor itself, but in its role as a proof of concept for one agent within a larger, orchestrated multi-agent systems. Orchestrated MAS offer a broader architectural vision in which specialised agents, such as Socratic guides, critical companions, and affective coaches, work in tandem, under pedagogical orchestration, to support diverse learner needs and trajectories. Unlike monolithic AI deployments, MAS are designed to be modular, work with the same knowledge, be pedagogically aligned, and epistemically diverse.

From this perspective, orchestrated MAS constitute a new learning infrastructure: a hybrid approach in which human and artificial agents collaboratively engage in sustained, reflective inquiry. This shift redefines educational roles. Educators become orchestrators of learning processes, curating agent constellations in line with instructional intent. Students become epistemic co-agents, responsible for navigating, evaluating, and synthesising input from a distributed field of human and non-human educators.

To realise this vision, four institutional commitments are essential. First, a pedagogical commitment to epistemic empowerment: AI systems must scaffold inquiry and foster agency, not automate cognition. Second, a technical and infrastructural commitment. We hope future multi-agent ecosystems to be developed as modular, extensible, and non-proprietary. Third, an ethical commitment to transparency, contestability, and human-centred design: orchestration layers must enable human oversight and value-sensitive alignment. And fourth, a commitment to overhaul curricula reflecting the changing educational landscape.

As generative systems become embedded in the everyday experience of learning, universities must foster not only AI literacy, but the human literacies that allow learners to thrive in hybrid epistemic communities. At their best, multi-agent systems enable new constellations of support, scalable, dialogic, and equitable. But their promise will only be fulfilled through intentional pedagogical design and sustained institutional commitments.

## 11. Declaration of generative AI and AI-assisted technologies in the writing process.

During the preparation of this work the authors used DeepL Write in order to find word- and sentence alternatives. After using this tool, the authors reviewed and edited the content as needed and takes full responsibility for the content of the published article.